\pdfoutput=1

\documentclass[11pt]{article}

\usepackage{acl}
\usepackage{latexsym,booktabs,array,adjustbox}
\usepackage{times}
\usepackage{latexsym}
\usepackage{booktabs}
\usepackage[T1]{fontenc}

\usepackage[utf8]{inputenc}

\usepackage{microtype}

\usepackage{inconsolata}

\usepackage{graphicx}

\usepackage{tikz}
\usetikzlibrary{trees}
\usepackage{lipsum}

\title{Language and Multimodal Models in Sports: A Survey of Datasets and Applications}

\author{Haotian Xia\textsuperscript{1},
Zhengbang Yang\textsuperscript{2},
Yun Zhao\textsuperscript{3},
Yuqing Wang\textsuperscript{4},
Jingxi Li\textsuperscript{1},
Rhys Tracy\textsuperscript{5}, \\
{\bf Zhuangdi Zhu\textsuperscript{2},
Yuan-fang Wang\textsuperscript{5},
Hanjie Chen\textsuperscript{6}$^*$,
Weining Shen\textsuperscript{1}}$^*$ \\
\textsuperscript{1}University of California, Irvine, CA, USA  \textsuperscript{2}George Mason University, VA, USA \\ \textsuperscript{3}Meta Platforms, Inc., CA, USA \textsuperscript{4}Stanford University, CA, USA\\
\textsuperscript{5}University of California, Santa Barbara, CA, USA\\
\textsuperscript{6} Rice University, TX, USA\\
  \texttt{\small\{xiah6, weinings\}@uci.edu},
  \texttt{\small\{zyang30, zzhu24\}@gmu.edu},
  \texttt{\small hanjie@rice.edu}
  }

\begin{document}
\maketitle
\def\thefootnote{*}\footnotetext{Corresponding authors.}\def\thefootnote{\arabic{footnote}}
\begin{abstract}
Recent integration of Natural Language Processing (NLP) and multimodal models has advanced the field of sports analytics. This survey presents a comprehensive review of the datasets and applications driving these innovations post-2020. We overviewed and categorized datasets into three primary types: language-based, multimodal, and convertible datasets. Language-based and multimodal datasets are for tasks involving text or multimodality (e.g., text, video, audio), respectively. 
Convertible datasets, initially single-modal (video), can be enriched with additional annotations, such as explanations of actions and video descriptions, to become multimodal, offering future potential for richer and more diverse applications. Our study highlights the contributions of these datasets to various applications, from improving fan experiences to supporting tactical analysis and medical diagnostics. We also discuss the challenges and future directions in dataset development, emphasizing the need for diverse, high-quality data to support real-time processing and personalized user experiences. This survey provides a foundational resource for researchers and practitioners aiming to leverage NLP and multimodal models in sports, offering insights into current trends and future opportunities in the field.
\end{abstract}

\section{Introduction}

The domain of sports, characterized by its dynamic nature and escalating popularity, stands as a testament to human endeavor and competition. As the world increasingly indulges in various sports, from global events like the FIFA World Cup to local leagues, the intersection of technology and sports has become a focal point of interest. In particular, the advancement of Natural Language Processing (NLP) has opened new avenues for enhancing user experiences~\citep{aug1-chen2022sporthesia}, performance analytics~\citep{per3-bera2023football}, and operational efficiencies~\citep{robinson2023evaluating}.

NLP has long been instrumental in sports analytics, providing key functionalities such as game summarization~\citep{new3-wang2022knowledge,huang-etal-2020-generating,belemkoabga-etal-2021-neural} and commentary generation~\citep{comment-gen1-qi2023goalmulti1, comment-gen2-kim2020automatic}. These applications have contributed to enhancing fan experiences and offering analytical support to coaches and players. With the advent of large language models (LLMs) such as GPT~\citep{openai2023gpt4}, Llama2~\citep{touvron2023llama}, and Gemini~\citep{team2023gemini}, the capabilities of NLP in sports have expanded tremendously. LLMs have introduced a new level of nuance and sophistication to these applications, enriching the depth of analytics. This evolution marks a significant milestone in the use of NLP technologies in sports, pushing the boundaries of what these tools can achieve in terms of data-driven insights and user interaction. For instance, the SNIL system~\citep{new5-cheng2024snil} uses LLMs to generate sports news articles that are more insightful and closely aligned with user-provided insights, improving the overall narrative quality and user satisfaction. Furthermore, these advancements open up new possibilities for the future, where NLP can not only refine existing applications but also innovate new ways to interpret, understand, and interact with sports data. For example, achieving expert-level performance on scenario-based questions in the SportsQA~\citep{xia2024sportqa} dataset means LLMs have sports understanding capacities close to those of sports experts. This could enhance LLM applications in areas such as AI refereeing, player mindset training, and tactics education.

Despite these advancements, there remains a notable gap in the literature: a comprehensive survey that encapsulates the breadth of datasets and applications driven by NLP and multimodal approaches within the sports domain. To the best of our knowledge, while there are surveys focusing on computer vision~\citep{zhao2023survey,wu2022survey} in sports, primarily in movement recognition and classification, the specific contributions of NLP and its integration with multimodal data have not been thoroughly explored.

\begin{figure*}[h]
	\centering
 \begin{adjustbox}{totalheight=0.35\textheight}
	\begin{tikzpicture}
		[ edge from parent fork right,
		grow = right,
		every node/.style       = {font=\fontsize{10}{10}\selectfont, rectangle, rounded corners, draw},
		level 1/.style={level distance= 11em},
		level 2/.style={level distance= 14em},
		level 3/.style={level distance= 18.9em},
		]
		\definecolor{myblue}{HTML}{d4edf4}
		\definecolor{myred}{HTML}{fdebeb}
		\definecolor{mygreen}{HTML}{eefced}
		\definecolor{mybluebox}{HTML}{04a4da}
		\definecolor{myredbox}{HTML}{f18382}
		\definecolor{mygreenbox}{HTML}{8fed8f}
		\tikzstyle{root} = [fill=gray!30, text=black, draw=gray,line width=1.5pt,text width=3cm]
		\tikzstyle{Node11} = [fill=myblue, text=black, draw=mybluebox,line width=1.5pt,text width=3.5cm]
		\tikzstyle{Node12} = [fill=myred, text=black, draw=myredbox,line width=1.5pt,text width=3cm]
		\tikzstyle{Node13} = [fill=mygreen, text=black, draw=mygreenbox,line width=1.5pt,text width=3cm]
		\tikzstyle{Node11child} = [fill=myblue, text=black, draw=mybluebox,line width=1pt,text width=3.5cm]
		\tikzstyle{Node12child} = [fill=myred, text=black, draw=myredbox,line width=1pt,text width=3cm]
		\tikzstyle{Node13child} = [fill=mygreen, text=black, draw=mygreenbox,line width=1pt,text width=5cm]
		\tikzstyle{Node21child} = [fill=myblue, text=black, draw=mybluebox,line width=0.5pt,text width=7cm]
		\tikzstyle{Node22child} = [fill=myred, text=black, draw=myredbox,line width=0.5pt,text width=7cm]
		\tikzstyle{Node23child} = [fill=mygreen, text=black, draw=mygreenbox,line width=0.5pt,text width=7m]

		\node [root]{Sports Dataset and Application}
		child[sibling distance=6em] { node [Node13] {Convertible Datasets}
			child [sibling distance=2.5em] { node[Node13child] {\citet{xia2022vren},\citet{con1-li2021multisports},\citet{con2-liu2020fsd}, \citet{con3-shao2020finegym}} 
				}
		}
		child [sibling distance=10em]{ node [Node12] {Multimodal Datasets}
			child [sibling distance=2.5em] { node[Node12child] {Video Understanding} 
				child { node[Node22child] {  ~\citet{li2024sports},~\citet{held2024x},\citet{biermann2021unified} } } }
			child [sibling distance=4em] { node[Node12child] { Video Augmentation} 
				child { node[Node22child] { \citet{aug1-chen2022sporthesia}  } } }
			child [sibling distance=2.8em] { node[Node12child] { Game Summarization } 
				child { node[Node22child] {  ~\citet{vsum1-sanabria2021hierarchical}, ~\citet{vsum2-qiu2024mmsum}, \citet{vsum4-narwal2023novel}, \citet{vsum3-khan2020content} } } }
		}
		child[sibling distance=17em] { node [Node11] {Language Model Datasets}
                    child [sibling distance=3em] { node[Node11child] { Medical } 
				child { node[Node21child] {~\citet{medical1-chen2024evaluating}, ~\citet{medical2-li2022automated}  } } }
                    child [sibling distance=3em] { 
            node[Node11child] {Education } 
				child { node[Node21child] {  ~\citet{edu-zeng2023textual} } } }
			child [sibling distance=3em] { node[Node11child] {Sports Understanding } 
				child { node[Node21child] { ~\citet{jardim2023qasports}, ~\citet{srivastava2023beyond}, ~\citet{xia2024sportqa} } } }
			child [sibling distance=3em] { node[Node11child] {Fan Engagement} 
				child { node[Node21child] {  \citet{qianying-etal-2020-liveqa}, \citet{fan1-ning2022sports}, \citet{fan2-h2021quantifying} } } }
			child [sibling distance=3.7em] { node[Node11child] {News Highlight Generation and Game Summarization} 
				child { node[Node21child] { \citet{huang-etal-2020-generating},\citet{new2-wang2021sportssum2},\citet{new3-wang2022knowledge}, \citet{thomson-etal-2020-sportsett}, etc.   } } }
			child [sibling distance=4em] { node[Node11child] { Named-Entity Recognition } 
				child { node[Node21child] {  ~\citet{name1-seti2020named},~\citet{name2-wijesinghe2022sinhala},~\citet{name3-liu2022named} } } }
			child [sibling distance=4em] { node[Node11child] {Hate Speech Detection} 
				child { node[Node21child] {~\citet{hate1-vujivcic2023approach}, ~\citet{hate2-toraman2022large}, ~\citet{hate3-romim2021hate}, ~\citet{hate4-aljarah2021intelligent}   } } }
                child [sibling distance=4em] { node[Node11child] {Game and Player Performance Prediction and Analysis} 
				child { node[Node21child] {  ~\citet{per2-oved2020predicting}, ~\citet{per3-bera2023football}, ~\citet{velichkov-etal-2019-deep} } } } 
		};
	\end{tikzpicture}
 \end{adjustbox}
	\caption{Taxonomy of Research on Sports Datasets and Corresponding Applications}
	\label{fig:sports_datasets}
\end{figure*}
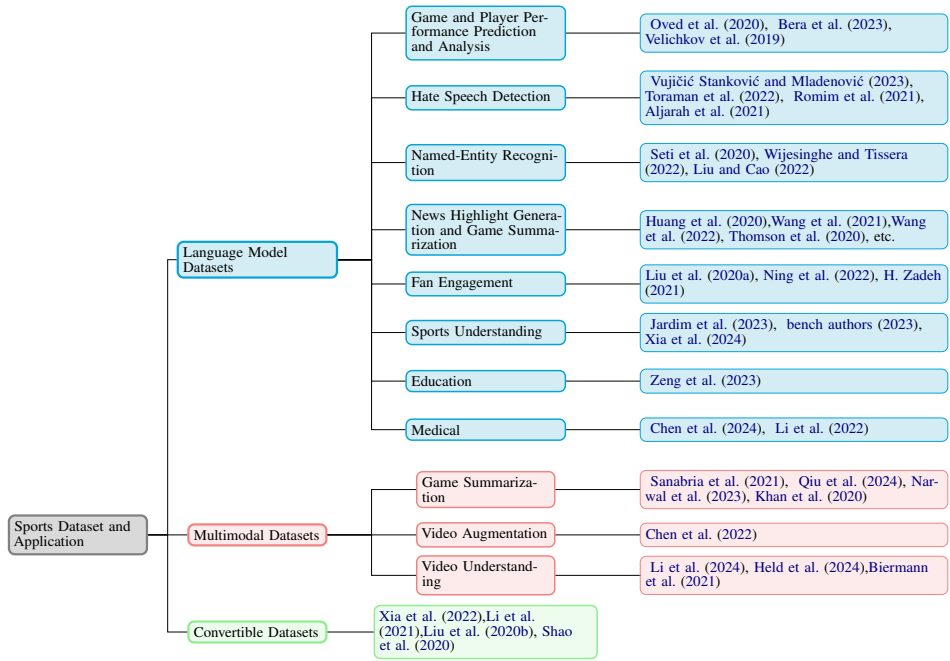

This review aims to bridge the gap by focusing on a taxonomy of datasets and applications that highlight the integration of NLP and multimodal models in sports. Particularly, our survey focuses on research and developments post-2020. Our study is organized into three principal categories: language-based datasets (\S\ref{sec2}), multimodal datasets (\S\ref{sec3}), and convertible datasets (\S\ref{sec4})—each forming a separate section of the survey. By ``convertible datasets'', we refer to datasets that can be transformed into multimodal datasets through future operations such as annotation. In each section, we systematically explore different applications, initially discussing their relevance to sports and subsequently detailing the specific datasets that support these applications. It should be noted that some datasets are versatile and can be applied to multiple applications. For simplicity, we categorize them based on their primary focus in the original papers, showing in Figure~\ref{fig:sports_datasets}. Additionally, an overview of the various types of datasets, including their size, source, and task type, is provided in Appendix~\ref{sec:appendix}. We also discuss future work and challenges in \S\ref{sec5}, outlining the potential directions and obstacles in advancing the integration of NLP and multimodal models in sports.
\begin{itemize}
    \item This is the first survey that systematically explores the integration of NLP and multimodal models in sports, setting a foundation for future academic and practical applications.
    \item The survey provides an extensive review of existing datasets and their applications, highlighting how advanced language models are currently utilized in the sports community.
    \item The survey offers a strategic roadmap for the future, suggesting how NLP and multimodal datasets can be effectively applied to sports applications to benefit the community.
\end{itemize}

By delving into these areas, this survey provides a vital resource for the sports community to harness the power of NLP and multimodal models in enhancing the sporting experience.

\section{Language-Based Datasets}
\label{sec2}
Language-based datasets are crucial in the evolving landscape of sports analytics and applications. By harnessing the power of NLP, these datasets enable deeper insights and more effective solutions across various domains, from medical applications and educational tools to enhancing fan engagement and understanding sports dynamics. This section introduces these datasets and illustrates their impact through specific applications.

\subsection{Game and Player Performance Prediction and Analysis}
Accurately predicting game outcomes and Analysis player performance is critical for teams, players, and coaches as it informs strategic planning, enhances decision-making, and optimizes performance on the field. 

The dataset by~\citet{per2-oved2020predicting} includes 1,337 pre-game interviews and corresponding performance metrics for key NBA players over 14 seasons, totaling 5,226 interview-metric pairs. This dataset aims to predict deviations from mean in-game actions, utilizing linguistic signals from interviews to forecast strategic choices, player behavior, and risk-related decisions. \citet{velichkov-etal-2019-deep} presents a dataset comprising 50 articles of pre-match interviews with athletes from individual sports like boxing, MMA, and tennis. This dataset also includes structured data such as sports rankings, ages, and previous match results, providing a comprehensive basis for predicting match outcomes based on the athletes' interviews and historical performance data. Substitution in soccer is essential for winning the game. ~\citeposs{per3-bera2023football} dataset combines past performance metrics and text converted from audio recordings of conversations between players and coaches for sentiment analysis to enhance substitution decisions in soccer.

\subsection{Hate Speech Detection}
In sports, where fans and athletes engage extensively online, hate speech can detract from the community experience and affect mental well-being. Effective detection and management of hate speech are essential to maintaining a supportive environment for all participants. This ensures that digital interactions around sports events remain respectful and inclusive, reflecting the true spirit of sportsmanship that underpins athletic competition.

The dataset provided by \citet{hate1-vujivcic2023approach} is designed to identify hate speech in sports-related communications on social media. The dataset includes data from multiple sources, focusing on comments published on popular entertainment and sports YouTube channels as well as sports news articles on Serbian news portals. \citet{hate2-toraman2022large} propose a dataset comprising 100,000 tweets across multiple domains, including sports, to test cross-domain model efficiency. \citet{hate3-romim2021hate} and \citet{hate4-aljarah2021intelligent} contribute by focusing on Bengali and Arabic languages, respectively, expanding the scope to include sports-related content within larger multilingual and multicultural datasets, with 30,000 user comments and 3,696 tweets analyzed, respectively.

\subsection{Named-Entity Recognition}
Named-Entity Recognition (NER) in sports efficiently extracts and categorizes key details like player names, team names, and event specifics from sports-related texts. This foundational technology supports complex systems beyond summarization, enhancing data analysis, fan engagement, and personalized content delivery by providing structured and accessible data. 

The dataset~\cite{name1-seti2020named} is crafted from major Chinese sports news websites and contains sports texts used to test a NER model. It features different types of entities, such as sporting event names, stadium names, team names, player names, and results. \citeposs{name2-wijesinghe2022sinhala} dataset is specifically built for Sinhala language sports-related content. It includes annotated data from 2,000 Sinhala e-news articles, focusing on named entities relevant to the sports domain such as tournament names, school names, and ground locations. ~\citet{name3-liu2022named} propose a dataset that includes detailed annotations across various entity types relevant to the sports domain. Specifically, the dataset features 12,802 sentences with nine different entity types, such as athletes, coaches, venues and sports items, which are crucial for enriching the content's contextual understanding and specificity.

\subsection{News Highlight Generation and Game Summarization}
In the sports media field, News Highlight Generation and Game Summarization serve critical roles by distilling extensive match details into digestible and engaging content. While both processes aim to encapsulate key moments, they do so with different focuses and objectives. News Highlight Generation captures the most impactful or emotionally charged events to enhance fan engagement,emphasizing dramatic moments, key plays, and exciting highlights. Conversely, Game Summarization provides a comprehensive and concise account of the game, emphasizing pivotal plays, outcomes, and statistics to give a complete and structured overview of the match. \citet{huang-etal-2020-generating} introduce the SportsSum dataset, which consists of 5,428 pairs of live commentaries and corresponding news articles derived from major soccer leagues. This dataset is specifically designed to address the challenges of summarizing sports games, where the narrative style of the source (commentaries) differs from that of the target summaries (news articles). Subsequent iteration, SportsSum2.0~\citep{new2-wang2021sportssum2}, refined SportsSum further by cleaning noisy data. Building on SportsSum2.0, the K-SportsSum~\citep{new3-wang2022knowledge} dataset not only increased the dataset to 7,854 commentary-news pairs but also enriched it with comprehensive background knowledge of teams and players, showcasing a pivotal development in bridging the knowledge gap essential for generating more informative sports news. 

\citeposs{thomson-etal-2020-sportsett} dataset is structured to facilitate better Natural Language Generation for sports by providing a comprehensive, multi-dimensional dataset of basketball statistics. It includes detailed game data from the 2014 to 2018 NBA seasons, organized in a way that allows for complex queries and supports a richer entity relationship graph, aimed at enhancing data-driven sports content creation. \citet{new4-10.1145/3606038.3616157} propose a dataset containing 143 annotated narratives from selected Premier League games, aimed at fine-tuning a Natural Language Inference model to ensure the factual accuracy of the news narratives. The dataset in the study~\citep{new5-cheng2024snil} uses a curated dataset of 120 sports news articles to demonstrate the application of large language models in sports journalism.

\subsection{Fan Engagement}
Fan engagement in sports is increasingly driven by data analytics, with organizations leveraging datasets to understand and enhance the fan experience. This subsection explores various datasets that can be used by language models to quantify and improve fan engagement using text analytics and sentiment analysis. 

\citet{qianying-etal-2020-liveqa} introduce the LiveQA dataset, derived from interactive content in NBA play-by-play live broadcasts, encompassing 117k multiple-choice questions from 1,670 games. This dataset captures dynamic fan interactions during live events, illustrating how engaging real-time content can boost fan participation and enhance the viewing experience. \citeposs{fan1-ning2022sports} dataset consists of 821 comments totaling 1,205 lines of text collected through social media surveys from basketball season ticket holders. This dataset was used to build models for text categorization and polarity assessments via sentiment analysis, providing insights into fan sentiments that can guide engagement strategies, \citet{fan2-h2021quantifying} leverage a dataset comprising 391,092 entries from Zhihu to analyze sports-related queries and discussions. This dataset facilitates a comprehensive understanding of the topics that dominate fan interactions and the types of information most sought after by different user groups within the community. This insight is essential for tailoring engagement and content strategies to meet the specific needs of sports enthusiasts effectively.

\subsection{Sports Understanding}
Sports Understanding is pivotal for the development of AI applications, such as AI referees, where the sophistication of model comprehension 2impacts their effectiveness and reliability. This subsection explores QA datasets that are instrumental in enhancing models' capabilities to interpret complex sports scenarios, rules, and statistics. These datasets train models to deliver a higher quality of content and interaction within the sports community, supporting both sports 
enthusiasts and professionals by providing deeper insights and a more intuitive understanding of sports through advanced technology. Such capabilities are essential for developing systems that can effectively analyze and respond to dynamic sports environments.

QASports~\citep{jardim2023qasports} is a context extractive question-answering dataset focusing on soccer, American football, and basketball. It primarily involves answering questions based on specifically provided paragraphs focused on fact-based questions. The BIG-bench sports understanding task~\citep{srivastava2023beyond}, is a subtask within the broader BIG-bench initiative designed to evaluate the sports-specific comprehension of large language models. This task presents 986 multiple-choice questions where models must discern plausible from implausible sports-related statements, requiring knowledge of athletes' names and actions typical in various sports. It challenges models to use higher-level reasoning to understand the nuances of sports contexts rather than merely recalling associations. 

The more challenging dataset, SportQA~\citep{xia2024sportqa}, is specially designed to test the sports understanding of large language models (LLMs) across three levels of difficulty, from basic historical facts and sports strategies and rules to advanced scenario-based reasoning. It contains 70,592 multiple-choice questions covering 35 different sports that challenge models to demonstrate deep comprehension of sports-related scenarios and facts. This dataset addresses the gap in sports-specific benchmarks for evaluating the nuanced understanding of sports in LLMs. 

\subsection{Medical Application}
Medical applications in sports, facilitated by advancements in NLP, are crucial for analyzing clinical texts to support injury diagnosis and treatment. These technologies are especially valuable in sports medicine for processing and interpreting radiological and surgical data, aiding in the timely and accurate medical care of athletes.

Knee Meniscal Tears Dataset~\citep{medical2-li2022automated} involves radiology and arthroscopy reports focused on knee meniscal tears, relevant for sports due to the frequency of knee injuries among athletes. The dataset includes 3,593 knee MRI reports from a single institution, providing a substantial basis for developing NLP models that automate the extraction and correlation of medical findings.

Managing osteoarthritis is important in sports, as the condition frequently results from sports-related injuries~\citep{medical1prove-bullock2020playing} and cans impact athletic performance and career longevity. The Osteoarthritis dataset~\citep{medical1-chen2024evaluating} consists of data from 80 real-world patients diagnosed with osteoarthritis, collected between April and October 2023. It includes comprehensive details like age, BMI, symptoms, and radiographic findings, derived from six well-established clinical guidelines. This dataset aims to evaluate the capability of large language models to generate personalized treatment plans based on evidence-based medicine.

\subsection{Educational Application}
Educational applications in sports increasingly utilize NLP to enhance teaching methodologies and student engagement. This involves analyzing educational content and interactions to provide personalized learning experiences and assess teaching effectiveness. \citet{edu-zeng2023textual} explores the effectiveness of digital teaching methods in sports education by utilizing a hybrid intelligent system that combines text and visual features for quality evaluation. The datasets used in this study are essential for assessing and enhancing the interactivity and educational value of sports training videos. Specifically, the Charades Dataset, consisting over 10,000 videos with 157 action types, and the Net-caption Dataset, with over 20,000 videos and 100,000 video segment-sentence pairs, are instrumental in developing models assessing digital sports education quality. 

\section{Multimodal Datasets}
\label{sec3}
Multimodal datasets integrate multiple types of data, such as text, video, and audio, to provide a richer and more comprehensive understanding of sports events. These datasets are crucial for developing advanced AI models that can interpret and analyze complex scenarios, offering deeper insights and more engaging experiences for both fans and professionals. In this section, we explore various applications of multimodal datasets in sports, including game summarization, video understanding, and video augmentation. Each subsection highlights specific datasets and their contributions to enhancing sports analytics and user experiences.

\subsection{Game Summarization}
Game Summarization leverages multimodal datasets to deliver concise and engaging sports content. These datasets encompass video, audio and textual data, providing a richer context for summarization technologies. These datasets enable the creation of video summaries that retain the original game's visual and auditory essence, enhancing the viewer's experience by highlighting crucial moments. The dataset~\citep{vsum1-sanabria2021hierarchical} consists of event data from 70 soccer matches across two European leagues, including a diverse range of event types like passes, shots, and fouls. Each match generates approximately 1,700 event instances which are used to train the summarization model. This comprehensive dataset allows for a nuanced analysis of game dynamics, aiming to use Multimodal Models to automate the generation of match highlights and summaries effectively. 

The MSMO dataset~\citep{vsum2-qiu2024mmsum} features 5,100 videos across various themes, with sports being one of the prominent categories. It provides both video and textual summaries, which are valuable for developing multimodal summarization techniques. This aspect makes it suited for applications that require combining visual content with descriptive text to enhance the viewer's understanding and engagement with sports events. \citet{vsum4-narwal2023novel} propose two multimodal datasets tailored for cricket video summarization: the DPCS (Delivery Play Cricket Sport) image dataset and the EXINP (Excited Interval Normal Play) Cricket audio dataset. The DPCS dataset contains 8,646 high-definition images divided into two classes, Delivery and Play, used for segmenting cricket videos. The EXINP dataset comprises 868 audio segments, categorized into Excited, Interval, and Normal Play, to identify key events. Together, these datasets utilize audio-visual cues for dynamic cricket video summarization. 

\citeposs{vsum3-khan2020content} include 104 broadcast sports videos, totaling over 230 hours of content. This dataset encompasses multiple sports types and includes additional elements like commercials, providing a comprehensive base for testing summarization models. It integrates audio and visual cues—such as crowd reactions and score changes—to identify key moments for video summarization, emphasizing its utility in creating enriched sports highlights that effectively capture the essence of events.

\subsection{Video Understanding}

Video understanding in sports utilizes multimodal models to interpret and analyze complex scenarios beyond the capabilities of traditional classification models. Unlike predefined task-based models, multimodal models can perform complex and open-ended tasks, such as detecting and describing whether a player has broken a rule, while CV classification models are more suited for identifying specific and well-defined movements performed by a player.

The Sports-QA dataset~\citep{li2024sports} comprises approximately 94,000 question-answer pairs across four types of questions derived from sports videos and professional action labels sourced from the MultiSports and FineGym datasets. Descriptive questions involve simple queries about the actions and events in the video. Temporal questions focus on the relationships among actions over time. Causal questions uncover the reasons or processes behind actions by exploring causal relationships. Counterfactual questions set hypothetical conditions not occurring in the video and query the expected outcomes based on these conditions. The dataset's complexity and diversity make it highly adaptable to multimodal settings, providing a robust foundation for testing the model's capabilities.

The X-VARS dataset~\citep{held2024x} enhances the evaluation of multimodal models with over 10,000 video clips and more than 22,000 video-question-answer triplets sourced from football games Annotated by over 70 experienced football referees. It includes complex, high-level questions that require identification and further explanation. Another contribution is the EIGD dataset\citep{biermann2021unified}. This dataset includes 125 minutes of handball and 125 minutes of soccer video. The handball videos are provided by the Deutsche Handball Liga and Kinexon, while the soccer videos are from publicly available broadcast recordings of the FIFA World Cup. The dataset features questions about low-level actions (free kicks, throw-ins) and high-level actions (passes, shots), enriching the training and evaluation of multimodal models in understanding various granularity levels in sports events.

\subsection{Video Augmentation}
Video augmentation in sports broadcasting harnesses multimodal datasets to enhance the viewer experience by integrating real-time data visualizations into live broadcasts. This not only enriches the visual presentation but also deepens fans' understanding of the game by providing insightful, data-driven analyses directly within the video feed. \citet{aug1-chen2022sporthesia} presents a dataset comprising 155 sports video clips paired with commentary text. This multimodal dataset is used to develop a system that augments sports videos by dynamically integrating visualized data from the commentaries into the videos. It serves as a foundation for developing practical applications of linking textual and visual data to enhance multimedia content.


\section{Convertible Datasets}
\label{sec4}
By "convertible datasets," we refer to datasets that are not yet multimodal but can be transformed into multimodal datasets through future operations such as annotation. We focus on the potential conversion to multimodal datasets rather than solely to language-based datasets because multimodal capabilities enable richer and more diverse applications in sports, enhancing the analysis and understanding of complex interactions within the data. The introduction of convertible datasets is particularly crucial given the relative scarcity of multimodal datasets compared to language-based datasets, underscoring the need to expand the availability and versatility of multimodal resources.

The VREN dataset~\citep{xia2022vren} introduces a generalized language for describing volleyball rallies through its dataset variable settings, meeting professional tactical statistics requirements. However, not all rallies are annotated. By annotating each rally and verifying the accuracy, this dataset can become multimodal. This allows large language models to generate similar descriptions, aiding both fans in understanding matches and professionals in real-time statistical analysis, while also enhancing model understanding of 
volleyball.

The MultiSports dataset~\citep{con1-li2021multisports} includes annotated action instances across basketball, volleyball, football, and aerobic gymnastics. The FineGym dataset~\citep{con3-shao2020finegym}provides hierarchical annotations for gymnastics, structured into events, sets, and elements, with both semantic and temporal annotations. The FSD-10 dataset~\citep{con2-liu2020fsd} contains video clips from figure skating competitions, annotated with action types, base values, grades of execution, and other metadata. By adding detailed textual descriptions, integrating sensor data, and including audio commentary, these datasets can become multimodal. We focus on these three datasets as representative examples to illustrate the potential of converting sports datasets into multimodal resources through annotation. Transforming these datasets will enable them to support advanced AI models by providing deeper contextual information, improving the granularity of real-time analysis, and facilitating the development of applications that help both fans and professionals comprehend complex sports 

\section{Future Work and Challenges}
\label{sec5}

In this section, we delve into the future directions and challenges associated with NLP and multimodal datasets in sports. As the first survey to systematically explore this integration, we emphasize the diverse applications of these technologies and the corresponding dataset requirements. While sections \ref{sec2} and \ref{sec3} categorized datasets based on their type and primary use, here we adopt a different approach to highlight the specific needs and potential impact on different backgrounds of people within the sports community. Our discussion is categorized into three main application areas: applications for enthusiasts, applications for professionals, and medical or rehabilitation applications. Each category not only highlights the current contributions but also outlines the challenges and future work needed to enhance these applications. Among all the challenges, data quality stands out as a critical factor that impacts the effectiveness of these applications.

\subsection{Applications for Enthusiasts}
Applications for sports enthusiasts primarily focus on enhancing fan engagement and understanding of the game. Key applications include highlight generation, news generation, and rule explanation augmentation. These applications are essential for maintaining and growing the fan base by providing richer, more engaging experiences.

\subsubsection{Highlight Generation and News Generation}
NLP and multimodal datasets have improved the generation of game highlights and sports news, making them more engaging and informative. Automated systems can now create concise and captivating highlights by analyzing game footage and generating corresponding textual summaries. These applications keep fans updated and involved, making sports more accessible and enjoyable.

\textbf{Challenges:} \textit{Dataset Diversity and Quality:} Ensuring datasets cover a wide range of sports and include diverse and high-quality video and textual data is crucial. Current datasets lack the breadth required to generalize across different sports and events.
\textit{Real-Time Processing:} The need for real-time highlight generation demands efficient processing capabilities and prompt data availability, which poses a significant challenge for existing datasets.
\textit{Fan Engagement Insights:} Understanding and predicting fan preferences through data-driven insights is essential. This requires datasets that not only capture game events but also fan reactions and interactions.
\subsubsection{Rule Explanation and Augmentation}
Enhancing fan understanding through applications that provide detailed rule explanations and augmented game insights in real-time can elevate the viewing experience. These applications make complex rules accessible to viewers, broadening the audience base by augmenting broadcasts with informative overlays and contextual details.

\textbf{Challenges:} \textit{Integration of Multimodal Data:} Combining video, audio, and textual data seamlessly to provide coherent and contextually accurate rule explanations requires advanced multimodal datasets.
\textit{Personalization:} Personalizing content based on individual viewer's knowledge level and preferences necessitates datasets that can support adaptive learning and recommendation systems.

\subsection{Applications for Professionals}
Applications designed for athletes, coaches, and sports analysts focus on real-time tactical analysis and performance monitoring. These applications are crucial for enhancing the competitive edge and performance of athletes and teams. Compared to applications for enthusiasts, professionals require more detailed, accurate, and high-quality data, as well as different focuses. For example, while both may benefit from highlight generation, professionals need highlights that focus on specific tactics and player performance under certain conditions, offering a more nuanced and detailed perspective.

\subsubsection{Tactical Analysis and Performance Monitoring}
NLP and multimodal datasets enable detailed analysis of player performance and team tactics, assisting coaches in making decisions during games. These applications include real-time data analytics, motion analysis, and tactical simulations. They are vital for optimizing performance and strategy.

\textbf{Challenges:} \textit{Real-Time Data Collection:} Accurate and timely data collection during games is critical. Current datasets need to improve in capturing real-time data efficiently.
\textit{High Precision and Reliability:} Applications for professionals demand high precision and reliability, which requires datasets with extensive and accurate annotations.
\textit{Integration of Various Data Types:} Integrating biometric, video, and textual data to provide comprehensive insights is complex and requires sophisticated datasets.
\subsubsection{Training and Simulation Tools}
Advanced training tools that simulate game scenarios and provide feedback on player performance can enhance training efficiency. These tools rely heavily on accurate datasets that reflect real-game conditions.

\textbf{Challenges:} \textit{Realism in Simulations:} Ensuring simulations are realistic and reflect actual game conditions requires highly detailed and contextually rich datasets.
\textit{Adaptive Learning:} Training tools need to adapt to the evolving skills of athletes, which requires datasets that can support continuous learning and adaptation.
\subsection{Medical and Rehabilitation Applications}
Applications in the medical domain focus on injury diagnosis, treatment planning, and psychological support for athletes. These applications ensure the health and longevity of athletes' careers.

\subsubsection{Injury Diagnosis and Treatment}
NLP and multimodal datasets play a crucial role in analyzing medical reports, imaging data, and clinical records to support injury diagnosis and treatment planning. Accurate diagnosis and effective treatment plans are essential for the quick recovery and sustained performance of athletes.

\textbf{Challenges:} \textit{Comprehensive Medical Datasets:} Creating comprehensive datasets that cover various types of sports injuries and include detailed medical histories, imaging data, and treatment outcomes is essential.
\textit{Data Privacy and Security:} Ensuring the privacy and security of medical data while making it accessible for research and application development is a significant challenge.
\textit{Expert Validation:} Collecting and validating medical data requires input from medical professionals to ensure accuracy and relevance, adding a layer of complexity to dataset creation.
\subsubsection{Psychological Support and Rehabilitation}
Providing psychological support and monitoring rehabilitation progress through NLP-driven applications can aid athletes in their recovery process. These applications benefit from datasets that include psychological assessments, treatment plans, and recovery progress.

\textbf{Challenges:} \textit{Holistic Data Integration:} Integrating psychological, physiological, and performance data to provide comprehensive support requires multifaceted datasets.
\textit{Personalization and Adaptation:} Datasets need to support personalized treatment and adaptation based on individual recovery trajectories.

\section{Conclusion}
In this paper, we systematically reviewed applications and various datasets in integrating NLP and multimodal models within the sports domain. We categorized datasets into language-based, multimodal, and convertible types, highlighting their applications and associated challenges. We discussed future work needed to enhance applications for enthusiasts, professionals, and medical or rehabilitation purposes. By addressing these challenges, we aim to inspire further advancements and practical applications that meet the evolving needs of the sports community.
\section*{Limitations}
While this survey provides a comprehensive review of the integration of NLP and multimodal models in sports, several limitations should be acknowledged.

Firstly, the categorization of datasets into specific applications is based on the primary focus of their respective papers. However, many datasets have the potential to be applied across multiple domains. We mainly categorize them based on their main focus in the original papers, aiming for clarity and precision, though this may occasionally reduce the comprehensiveness of our classification. Secondly, the survey focuses on datasets and developments post-2020, potentially excluding valuable earlier contributions that could still be relevant. This temporal limitation is intended to highlight recent advancements but may inadvertently overlook foundational work that laid the groundwork for current innovations. Lastly, while we have highlighted key datasets and applications, the rapid evolution of both sports and technology means that new datasets and models are continuously being developed. Therefore, this survey represents a snapshot in time and should be updated regularly to reflect ongoing progress in the field.

\bibliography{anthology,custom}

\begin{thebibliography}{47}
\providecommand{\natexlab}[1]{#1}

\bibitem[{Aljarah et~al.(2021)Aljarah, Habib, Hijazi, Faris, Qaddoura, Hammo, Abushariah, and Alfawareh}]{hate4-aljarah2021intelligent}
Ibrahim Aljarah, Maria Habib, Neveen Hijazi, Hossam Faris, Raneem Qaddoura, Bassam Hammo, Mohammad Abushariah, and Mohammad Alfawareh. 2021.
\newblock Intelligent detection of hate speech in arabic social network: A machine learning approach.
\newblock \emph{Journal of Information Science}, 47(4):483--501.

\bibitem[{Belemkoabga et~al.(2021)Belemkoabga, Bossard, Essa, Rodrigues, and Sylla}]{belemkoabga-etal-2021-neural}
David~St{\'e}phane Belemkoabga, Aur{\'e}lien Bossard, Abdallah Essa, Christophe Rodrigues, and K{\'e}vin Sylla. 2021.
\newblock \href {https://aclanthology.org/2021.ranlp-1.18} {Neural network-based generation of sport summaries: A preliminary study}.
\newblock In \emph{Proceedings of the International Conference on Recent Advances in Natural Language Processing (RANLP 2021)}, pages 147--154, Held Online. INCOMA Ltd.

\bibitem[{bench authors(2023)}]{srivastava2023beyond}
BIG bench authors. 2023.
\newblock \href {https://openreview.net/forum?id=uyTL5Bvosj} {Beyond the imitation game: Quantifying and extrapolating the capabilities of language models}.
\newblock \emph{Transactions on Machine Learning Research}.

\bibitem[{Bera et~al.(2023)Bera, Joshi, and Nair}]{per3-bera2023football}
Aditya Bera, Savio Joshi, and Akhil~M Nair. 2023.
\newblock Football player substitution analysis using nlp and survival analysis.
\newblock In \emph{2023 5th International Conference on Smart Systems and Inventive Technology (ICSSIT)}, pages 753--757. IEEE.

\bibitem[{Biermann et~al.(2021)Biermann, Theiner, Bassek, Raabe, Memmert, and Ewerth}]{biermann2021unified}
Henrik Biermann, Jonas Theiner, Manuel Bassek, Dominik Raabe, Daniel Memmert, and Ralph Ewerth. 2021.
\newblock A unified taxonomy and multimodal dataset for events in invasion games.
\newblock In \emph{Proceedings of the 4th International Workshop on Multimedia Content Analysis in Sports}, pages 1--10.

\bibitem[{Bullock et~al.(2020)Bullock, Collins, Peirce, Arden, and Filbay}]{medical1prove-bullock2020playing}
Garrett~S Bullock, Gary~S Collins, Nick Peirce, Nigel~K Arden, and Stephanie~R Filbay. 2020.
\newblock Playing sport injured is associated with osteoarthritis, joint pain and worse health-related quality of life: a cross-sectional study.
\newblock \emph{BMC musculoskeletal disorders}, 21:1--11.

\bibitem[{Chen et~al.(2024)Chen, You, Wang, Liu, Fu, Xu, Zhang, Chen, and Li}]{medical1-chen2024evaluating}
Xi~Chen, MingKe You, Li~Wang, WeiZhi Liu, Yu~Fu, Jie Xu, Shaoting Zhang, Gang Chen, and Jian Li. 2024.
\newblock Evaluating and enhancing large language models performance in domain-specific medicine: Osteoarthritis management with docoa.
\newblock \emph{arXiv preprint arXiv:2401.12998}.

\bibitem[{Chen et~al.(2022)Chen, Yang, Xie, Beyer, Xia, Wu, and Pfister}]{aug1-chen2022sporthesia}
Zhutian Chen, Qisen Yang, Xiao Xie, Johanna Beyer, Haijun Xia, Yingcai Wu, and Hanspeter Pfister. 2022.
\newblock Sporthesia: Augmenting sports videos using natural language.
\newblock \emph{IEEE transactions on visualization and computer graphics}, 29(1):918--928.

\bibitem[{Cheng et~al.(2024)Cheng, Deng, Xie, Qiu, Xu, and Wu}]{new5-cheng2024snil}
Liqi Cheng, Dazhen Deng, Xiao Xie, Rihong Qiu, Mingliang Xu, and Yingcai Wu. 2024.
\newblock Snil: Generating sports news from insights with large language models.
\newblock \emph{IEEE Transactions on Visualization and Computer Graphics}.

\bibitem[{H.~Zadeh(2021)}]{fan2-h2021quantifying}
Amir H.~Zadeh. 2021.
\newblock Quantifying fan engagement in sports using text analytics.
\newblock \emph{Journal of Data, Information and Management}, 3(3):197--208.

\bibitem[{Held et~al.(2024)Held, Itani, Cioppa, Giancola, Ghanem, and Van~Droogenbroeck}]{held2024x}
Jan Held, Hani Itani, Anthony Cioppa, Silvio Giancola, Bernard Ghanem, and Marc Van~Droogenbroeck. 2024.
\newblock X-vars: Introducing explainability in football refereeing with multi-modal large language models.
\newblock In \emph{Proceedings of the IEEE/CVF Conference on Computer Vision and Pattern Recognition}, pages 3267--3279.

\bibitem[{Huang et~al.(2020)Huang, Li, and Chang}]{huang-etal-2020-generating}
Kuan-Hao Huang, Chen Li, and Kai-Wei Chang. 2020.
\newblock \href {https://aclanthology.org/2020.aacl-main.61} {Generating sports news from live commentary: A {C}hinese dataset for sports game summarization}.
\newblock In \emph{Proceedings of the 1st Conference of the Asia-Pacific Chapter of the Association for Computational Linguistics and the 10th International Joint Conference on Natural Language Processing}, pages 609--615, Suzhou, China. Association for Computational Linguistics.

\bibitem[{Jardim et~al.(2023)Jardim, Moraes, and Aguiar}]{jardim2023qasports}
Pedro~Calciolari Jardim, Leonardo Mauro~Pereira Moraes, and Cristina~Dutra Aguiar. 2023.
\newblock Qasports: A question answering dataset about sports.
\newblock In \emph{Anais do V Dataset Showcase Workshop}, pages 1--12. SBC.

\bibitem[{Khan et~al.(2020)Khan, Shao, Ali, and Tumrani}]{vsum3-khan2020content}
Abdullah~Aman Khan, Jie Shao, Waqar Ali, and Saifullah Tumrani. 2020.
\newblock Content-aware summarization of broadcast sports videos: an audio--visual feature extraction approach.
\newblock \emph{Neural Processing Letters}, 52(3):1945--1968.

\bibitem[{Kim and Choi(2020)}]{comment-gen2-kim2020automatic}
Byeong~Jo Kim and Yong~Suk Choi. 2020.
\newblock Automatic baseball commentary generation using deep learning.
\newblock In \emph{Proceedings of the 35th Annual ACM Symposium on Applied Computing}, pages 1056--1065.

\bibitem[{Li et~al.(2024)Li, Deng, Ke, Liu, Rahmani, Guo, Schiele, and Chen}]{li2024sports}
Haopeng Li, Andong Deng, Qiuhong Ke, Jun Liu, Hossein Rahmani, Yulan Guo, Bernt Schiele, and Chen Chen. 2024.
\newblock Sports-qa: A large-scale video question answering benchmark for complex and professional sports.
\newblock \emph{arXiv preprint arXiv:2401.01505}.

\bibitem[{Li et~al.(2022)Li, Deng, Chang, Kalpathy-Cramer, and Huang}]{medical2-li2022automated}
Matthew~D Li, Francis Deng, Ken Chang, Jayashree Kalpathy-Cramer, and Ambrose~J Huang. 2022.
\newblock Automated radiology-arthroscopy correlation of knee meniscal tears using natural language processing algorithms.
\newblock \emph{Academic radiology}, 29(4):479--487.

\bibitem[{Li et~al.(2021)Li, Chen, He, Wang, Wu, and Wang}]{con1-li2021multisports}
Yixuan Li, Lei Chen, Runyu He, Zhenzhi Wang, Gangshan Wu, and Limin Wang. 2021.
\newblock Multisports: A multi-person video dataset of spatio-temporally localized sports actions.
\newblock In \emph{Proceedings of the IEEE/CVF International Conference on Computer Vision}, pages 13536--13545.

\bibitem[{Liu and Cao(2022)}]{name3-liu2022named}
Pingshan Liu and Yuan Cao. 2022.
\newblock A named entity recognition method for chinese winter sports news based on roberta-wwm.
\newblock In \emph{2022 3rd International Conference on Big Data, Artificial Intelligence and Internet of Things Engineering (ICBAIE)}, pages 785--790. IEEE.

\bibitem[{Liu et~al.(2020{\natexlab{a}})Liu, Jiang, Wang, and Li}]{qianying-etal-2020-liveqa}
Qianying Liu, Sicong Jiang, Yizhong Wang, and Sujian Li. 2020{\natexlab{a}}.
\newblock \href {https://aclanthology.org/2020.ccl-1.98} {{L}ive{QA}: A question answering dataset over sports live}.
\newblock In \emph{Proceedings of the 19th Chinese National Conference on Computational Linguistics}, pages 1057--1067, Haikou, China. Chinese Information Processing Society of China.

\bibitem[{Liu et~al.(2020{\natexlab{b}})Liu, Liu, Huang, Feng, Hu, Jiang, Zhang, Liu, and Qiao}]{con2-liu2020fsd}
Shenlan Liu, Xiang Liu, Gao Huang, Lin Feng, Lianyu Hu, Dong Jiang, Aibin Zhang, Yang Liu, and Hong Qiao. 2020{\natexlab{b}}.
\newblock Fsd-10: a dataset for competitive sports content analysis.
\newblock \emph{arXiv preprint arXiv:2002.03312}.

\bibitem[{Narwal et~al.(2023)Narwal, Duhan, and Bhatia}]{vsum4-narwal2023novel}
Pulkit Narwal, Neelam Duhan, and Komal~Kumar Bhatia. 2023.
\newblock A novel multi-modal neural network approach for dynamic and generic sports video summarization.
\newblock \emph{Engineering Applications of Artificial Intelligence}, 126:106964.

\bibitem[{Ning et~al.(2022)Ning, Xu, Gao, Yang, and Wang}]{fan1-ning2022sports}
Chuanlin Ning, Jian Xu, Hao Gao, Xi~Yang, and Tianyi Wang. 2022.
\newblock Sports information needs in chinese online q\&a community: topic mining based on bert.
\newblock \emph{Applied Sciences}, 12(9):4784.

\bibitem[{OpenAI(2023)}]{openai2023gpt4}
OpenAI. 2023.
\newblock \href {https://arxiv.org/abs/2303.08774} {Gpt-4 technical report}.
\newblock \emph{Preprint}, arXiv:2303.08774.

\bibitem[{Oved et~al.(2020)Oved, Feder, and Reichart}]{per2-oved2020predicting}
Nadav Oved, Amir Feder, and Roi Reichart. 2020.
\newblock Predicting in-game actions from interviews of nba players.
\newblock \emph{Computational Linguistics}, 46(3):667--712.

\bibitem[{Qi et~al.(2023)Qi, Yu, Tu, Gao, Xu, Guan, Wang, Xu, Hou, Li et~al.}]{comment-gen1-qi2023goalmulti1}
Ji~Qi, Jifan Yu, Teng Tu, Kunyu Gao, Yifan Xu, Xinyu Guan, Xiaozhi Wang, Bin Xu, Lei Hou, Juanzi Li, et~al. 2023.
\newblock Goal: A challenging knowledge-grounded video captioning benchmark for real-time soccer commentary generation.
\newblock In \emph{Proceedings of the 32nd ACM International Conference on Information and Knowledge Management}, pages 5391--5395.

\bibitem[{Qiu et~al.(2024)Qiu, Zhu, Han, Kumar, Mittal, Jin, Yang, Li, Wang, Zhao et~al.}]{vsum2-qiu2024mmsum}
Jielin Qiu, Jiacheng Zhu, William Han, Aditesh Kumar, Karthik Mittal, Claire Jin, Zhengyuan Yang, Linjie Li, Jianfeng Wang, Ding Zhao, et~al. 2024.
\newblock Mmsum: A dataset for multimodal summarization and thumbnail generation of videos.
\newblock In \emph{Proceedings of the IEEE/CVF Conference on Computer Vision and Pattern Recognition}, pages 21909--21921.

\bibitem[{Robinson(2023)}]{robinson2023evaluating}
Drew Robinson. 2023.
\newblock Evaluating the potential of ai in sports consulting: Investigating chatgpt-4's ability to consult an mlb team.

\bibitem[{Romim et~al.(2021)Romim, Ahmed, Talukder, and Saiful~Islam}]{hate3-romim2021hate}
Nauros Romim, Mosahed Ahmed, Hriteshwar Talukder, and Md~Saiful~Islam. 2021.
\newblock Hate speech detection in the bengali language: A dataset and its baseline evaluation.
\newblock In \emph{Proceedings of International Joint Conference on Advances in Computational Intelligence: IJCACI 2020}, pages 457--468. Springer.

\bibitem[{Sanabria et~al.(2021)Sanabria, Precioso, and Menguy}]{vsum1-sanabria2021hierarchical}
Melissa Sanabria, Fr{\'e}d{\'e}ric Precioso, and Thomas Menguy. 2021.
\newblock Hierarchical multimodal attention for deep video summarization.
\newblock In \emph{2020 25th International conference on pattern recognition (ICPR)}, pages 7977--7984. IEEE.

\bibitem[{Sarfati et~al.(2023)Sarfati, Yerushalmy, Chertok, and Keller}]{new4-10.1145/3606038.3616157}
Noah Sarfati, Ido Yerushalmy, Michael Chertok, and Yosi Keller. 2023.
\newblock \href {https://doi.org/10.1145/3606038.3616157} {Generating factually consistent sport highlights narrations}.
\newblock In \emph{Proceedings of the 6th International Workshop on Multimedia Content Analysis in Sports}, MMSports '23, page 15–22, New York, NY, USA. Association for Computing Machinery.

\bibitem[{Seti et~al.(2020)Seti, Wumaier, Yibulayin, Paerhati, Wang, and Saimaiti}]{name1-seti2020named}
Xieraili Seti, Aishan Wumaier, Turgen Yibulayin, Diliyaer Paerhati, Lulu Wang, and Alimu Saimaiti. 2020.
\newblock Named-entity recognition in sports field based on a character-level graph convolutional network.
\newblock \emph{Information}, 11(1):30.

\bibitem[{Shao et~al.(2020)Shao, Zhao, Dai, and Lin}]{con3-shao2020finegym}
Dian Shao, Yue Zhao, Bo~Dai, and Dahua Lin. 2020.
\newblock Finegym: A hierarchical video dataset for fine-grained action understanding.
\newblock In \emph{Proceedings of the IEEE/CVF conference on computer vision and pattern recognition}, pages 2616--2625.

\bibitem[{Team et~al.(2023)Team, Anil, Borgeaud, Wu, Alayrac, Yu, Soricut, Schalkwyk, Dai, Hauth et~al.}]{team2023gemini}
Gemini Team, Rohan Anil, Sebastian Borgeaud, Yonghui Wu, Jean-Baptiste Alayrac, Jiahui Yu, Radu Soricut, Johan Schalkwyk, Andrew~M Dai, Anja Hauth, et~al. 2023.
\newblock Gemini: a family of highly capable multimodal models.
\newblock \emph{arXiv preprint arXiv:2312.11805}.

\bibitem[{Thomson et~al.(2020)Thomson, Reiter, and Sripada}]{thomson-etal-2020-sportsett}
Craig Thomson, Ehud Reiter, and Somayajulu Sripada. 2020.
\newblock \href {https://aclanthology.org/2020.intellang-1.4} {{S}port{S}ett:basketball - a robust and maintainable data-set for natural language generation}.
\newblock In \emph{Proceedings of the Workshop on Intelligent Information Processing and Natural Language Generation}, pages 32--40, Santiago de Compostela, Spain. Association for Computational Lingustics.

\bibitem[{Toraman et~al.(2022)Toraman, {\c{S}}ahinu{\c{c}}, and Yilmaz}]{hate2-toraman2022large}
Cagri Toraman, Furkan {\c{S}}ahinu{\c{c}}, and Eyup~Halit Yilmaz. 2022.
\newblock Large-scale hate speech detection with cross-domain transfer.
\newblock \emph{arXiv preprint arXiv:2203.01111}.

\bibitem[{Touvron et~al.(2023)Touvron, Martin, Stone, Albert, Almahairi, Babaei, Bashlykov, Batra, Bhargava, Bhosale et~al.}]{touvron2023llama}
Hugo Touvron, Louis Martin, Kevin Stone, Peter Albert, Amjad Almahairi, Yasmine Babaei, Nikolay Bashlykov, Soumya Batra, Prajjwal Bhargava, Shruti Bhosale, et~al. 2023.
\newblock Llama 2: Open foundation and fine-tuned chat models.
\newblock \emph{arXiv preprint arXiv:2307.09288}.

\bibitem[{Velichkov et~al.(2019)Velichkov, Koychev, and Boytcheva}]{velichkov-etal-2019-deep}
Boris Velichkov, Ivan Koychev, and Svetla Boytcheva. 2019.
\newblock \href {https://doi.org/10.26615/978-954-452-056-4_142} {Deep learning contextual models for prediction of sport event outcome from sportsman{'}s interviews}.
\newblock In \emph{Proceedings of the International Conference on Recent Advances in Natural Language Processing (RANLP 2019)}, pages 1240--1246, Varna, Bulgaria. INCOMA Ltd.

\bibitem[{Vuji{\v{c}}i{\'c}~Stankovi{\'c} and Mladenovi{\'c}(2023)}]{hate1-vujivcic2023approach}
Sta{\v{s}}a Vuji{\v{c}}i{\'c}~Stankovi{\'c} and Miljana Mladenovi{\'c}. 2023.
\newblock An approach to automatic classification of hate speech in sports domain on social media.
\newblock \emph{Journal of Big Data}, 10(1):109.

\bibitem[{Wang et~al.(2021)Wang, Li, Yang, Qu, Chen, Liu, and Hu}]{new2-wang2021sportssum2}
Jiaan Wang, Zhixu Li, Qiang Yang, Jianfeng Qu, Zhigang Chen, Qingsheng Liu, and Guoping Hu. 2021.
\newblock Sportssum2. 0: Generating high-quality sports news from live text commentary.
\newblock In \emph{Proceedings of the 30th ACM International Conference on Information \& Knowledge Management}, pages 3463--3467.

\bibitem[{Wang et~al.(2022)Wang, Li, Zhang, Zheng, Qu, Liu, Zhao, and Chen}]{new3-wang2022knowledge}
Jiaan Wang, Zhixu Li, Tingyi Zhang, Duo Zheng, Jianfeng Qu, An~Liu, Lei Zhao, and Zhigang Chen. 2022.
\newblock Knowledge enhanced sports game summarization.
\newblock In \emph{Proceedings of the Fifteenth ACM International Conference on Web Search and Data Mining}, pages 1045--1053.

\bibitem[{Wijesinghe and Tissera(2022)}]{name2-wijesinghe2022sinhala}
WMSK Wijesinghe and Muditha Tissera. 2022.
\newblock Sinhala named entity recognition model: Domain-specific classes in sports.
\newblock In \emph{2022 4th International Conference on Advancements in Computing (ICAC)}, pages 138--143. IEEE.

\bibitem[{Wu et~al.(2022)Wu, Wang, Bian, Ding, Lu, Cheng, Dou, and Xiong}]{wu2022survey}
Fei Wu, Qingzhong Wang, Jiang Bian, Ning Ding, Feixiang Lu, Jun Cheng, Dejing Dou, and Haoyi Xiong. 2022.
\newblock A survey on video action recognition in sports: Datasets, methods and applications.
\newblock \emph{IEEE Transactions on Multimedia}.

\bibitem[{Xia et~al.(2022)Xia, Tracy, Zhao, Fraisse, Wang, and Petzold}]{xia2022vren}
Haotian Xia, Rhys Tracy, Yun Zhao, Erwan Fraisse, Yuan-Fang Wang, and Linda Petzold. 2022.
\newblock Vren: volleyball rally dataset with expression notation language.
\newblock In \emph{2022 IEEE International Conference on Knowledge Graph (ICKG)}, pages 337--346. IEEE.

\bibitem[{Xia et~al.(2024)Xia, Yang, Wang, Tracy, Zhao, Huang, Chen, Zhu, Wang, and Shen}]{xia2024sportqa}
Haotian Xia, Zhengbang Yang, Yuqing Wang, Rhys Tracy, Yun Zhao, Dongdong Huang, Zezhi Chen, Yan Zhu, Yuan-fang Wang, and Weining Shen. 2024.
\newblock Sportqa: A benchmark for sports understanding in large language models.
\newblock \emph{arXiv preprint arXiv:2402.15862}.

\bibitem[{Zeng et~al.(2023)Zeng, Zhao, and Wen}]{edu-zeng2023textual}
Boyi Zeng, Jun Zhao, and Shantian Wen. 2023.
\newblock A textual and visual features-jointly driven hybrid intelligent system for digital physical education teaching quality evaluation.
\newblock \emph{Mathematical Biosciences and Engineering}, 20(8):13581--13601.

\bibitem[{Zhao et~al.(2023)Zhao, Chai, Hao, Hu, Wang, Cao, Song, Hwang, and Wang}]{zhao2023survey}
Zhonghan Zhao, Wenhao Chai, Shengyu Hao, Wenhao Hu, Guanhong Wang, Shidong Cao, Mingli Song, Jenq-Neng Hwang, and Gaoang Wang. 2023.
\newblock A survey of deep learning in sports applications: Perception, comprehension, and decision.
\newblock \emph{arXiv preprint arXiv:2307.03353}.

\end{thebibliography}

\appendix
\section{Appendix}
\label{sec:appendix}
Table~\ref{tab:02} shows the comprehensive overview of language-based, multimodal, and convertible datasets for sports applications.

\begin{table*}[h!]
	\centering
  \begin{adjustbox}{totalheight=0.45\textheight}
	\small
	\begin{tabular}{@{}m{4cm}m{4cm}m{5cm}m{6cm}@{}}
		\toprule
		Dataset & Size  & Data Source & Task Type  \\
		\midrule
		\multicolumn{4}{c}{Language-Based Dataset} \\ \hline
		\citet{medical2-li2022automated} & 3,593  & MRI reports & Detection  \\
		\citet{medical1-chen2024evaluating} &  80 & Clinical guidelines & Open-ended-QA \\
		\citet{edu-zeng2023textual}&110,000   & Combination of different datasets & Feature Extraction \& Evaluation+Description Generation \\
		\citet{hate1-vujivcic2023approach} & 180,785 & YouTube and News Portals comments & Classification(yes/no) \\
		\citet{hate2-toraman2022large} & 200,000 & Tweets & Classification \\
		\citet{hate3-romim2021hate} & 30,000 & YouTube and Facebook comments & Classification \\
		\citet{hate4-aljarah2021intelligent} & 3,696  & Arabic tweets & Classification \\
		\citet{name1-seti2020named} & 97,211 & Combination of existing datasets & Named-entity Recognition  \\
		\citet{name2-wijesinghe2022sinhala} & 99,972 & Sinhala sport news & Named-entity Recognition  \\
		\citet{name3-liu2022named} & N/A  & News portals & Named-entity Recognition \\
		\citet{huang-etal-2020-generating} & 5,428 & Sina Sports Live & Summarization Generation  \\
		\citet{new2-wang2021sportssum2} & 5,402  & Chinese sports websites & News Generation  \\
		\citet{new3-wang2022knowledge} & 7,854 & Comments from Sina Sports Live & Summarization Generation \\
		\citet{thomson-etal-2020-sportsett} & N/A & Rotowire, basketball-reference and Wikipedia & Summarization Generation\\
		\citet{new4-10.1145/3606038.3616157} & N/A & Narrative Generation \\
		\citet{new5-cheng2024snil} &N/A & NBA Games & Narrative Generation  \\
		\citet{qianying-etal-2020-liveqa} &  117,000 & Hupu Section of NBA Games & Multiple-Choice-QA \\
		\citet{fan1-ning2022sports} & 391,092 & Zhihu Section of Sports & Used to Study the Sport Community  \\
		\citet{fan2-h2021quantifying} & 821 & Survey of basketball Season Ticket Holders & Text Mining and Sentiment Analysis  \\
  \citet{per2-oved2020predicting} & 5,226 & Interviews with NBA Players & Player Performance Analysis \\
  \citet{per3-bera2023football} & N/A & Audio Recordings and Past Performance Metrics & Player Performance Analysis   \\
  \citet{velichkov-etal-2019-deep} & N/A & Interviews with Athletes & Game Outcome Prediction   \\
		&       &       &        \\ \hline
		\multicolumn{4}{c}{Multimodal Dataset} \\\hline
	\citet{vsum1-sanabria2021hierarchical}     & N/A &Human Observers in Matches     & Summarization      \\
		\citet{vsum2-qiu2024mmsum}    & 5,100     & Videos from YouTube, Annotations by Human Experts     & Summarization     \\
		\citet{vsum3-khan2020content}     & 104     & Multiple Video-sharing Platforms    & Highlight generation     \\
		\citet{vsum4-narwal2023novel}    & 9514    & Cricket Tournaments     & Video Segmentation+Audio Classification     \\
    \citet{jardim2023qasports} & More than 1.5 million & Fandom sports wikis & Open-ended-QA\\
    \citet{srivastava2023beyond} & 986 & Generated by the Author & Classification\\
    \citet{xia2024sportqa} & 70,592 & Existing QA datasets & Multiple-Choice-QA\\
  \hline
		\multicolumn{4}{c}{Convertible Datasets } \\\hline
		\citet{xia2022vren}     & 1,632     & NCAA men's Volleyball Games     & Prediction and Analyzation     \\
		\citet{con1-li2021multisports}    & 3,200    & Multiple Competition Records     & Classification      \\
		\citet{con2-liu2020fsd}     & 1,484     & Figure Skating Championships     & Multiple tasks    \\
		\citet{con3-shao2020finegym}    & N/A    & Gymnastics Competitions     & Multiple Tasks     \\
		\bottomrule
	\end{tabular}%
 \end{adjustbox}
 \caption{Overview of Language-Based, Multimodal, and Convertible Datasets for Sports Applications. N/A indicates that the specific data size was not mentioned in the original papers.}
	\label{tab:02}%
\end{table*}%

\end{document}